\documentclass[11pt,a4paper]{article}
\usepackage[hyperref]{emnlp2020}
\usepackage{times}
\usepackage{latexsym}

\usepackage{microtype}
\usepackage{booktabs}
\usepackage{float}
\usepackage{amsmath}
\usepackage{graphicx}
\usepackage{multirow}
\usepackage{caption}
\usepackage{xcolor}
\newenvironment{fontppl}{\fontfamily{ppl}\selectfont}{\par} % Palatino
\aclfinalcopy % Uncomment this line for the final submission
\usepackage{epsfig}
\usepackage{graphicx}
\usepackage{wrapfig}
\usepackage{subfig}
\usepackage[belowskip=0pt,aboveskip=0pt,font=small]{caption}
\usepackage{todonotes}
\presetkeys{todonotes}{inline}{}

\title{An Empirical Investigation Towards Efficient Multi-Domain Language Model Pre-training}

 \author{Kristjan Arumae, Qing Sun, \& Parminder Bhatia \\
  Amazon\\
  Seattle, USA \\
  \texttt{\{arumae, qinsun, parmib\}@amazon.com}}
   
\date{}

\begin{document}
\maketitle
\begin{abstract}

% \parry{Abstract alignment with the narrative}
Pre-training large language models has become a standard in the natural language processing community.
Such models are pre-trained on generic data (e.g. BookCorpus and English Wikipedia) and often fine-tuned on tasks in the same domain.
%  Out of domain tasks such as clinical named entity recognition and relation extraction, however, achieve state-of-the-art performance using additional in domain pre-training.
However, in order to achieve state-of-the-art performance on out of domain tasks such as clinical named entity recognition and relation extraction, additional in domain pre-training is required.
In practice, staged multi-domain pre-training presents performance deterioration in the form of \textit{catastrophic forgetting} (CF) when evaluated on a generic benchmark such as GLUE.
In this paper we conduct an empirical investigation into known methods to mitigate CF.
We find that elastic weight consolidation provides best overall scores yielding only a $0.33\%$ drop in performance across seven generic tasks while remaining competitive in bio-medical tasks.
Furthermore, we explore gradient and latent clustering based data selection techniques to improve coverage when using elastic weight consolidation and experience replay methods.

% With these methods we explore how best to reduce the performance gap introduced by domain shifts, when evaluated using domain specific fine-tuning tasks.
% Finally, we perform layer-wise probing of our models to get a sense of understanding of how to best transfer knowledge across domains in a continual learning setting.

% We find that with experience replay and elastic weight consolidation a heuristic based data sampling yields more robust models with less data, and a layer-wise understanding of deep models further improves performance.
\end{abstract}

\section{Introduction}

Transformer \citep{NIPS2017_7181} based language modeling has taken over many previous pre-training and initialization approaches \citep{devlin2018bert, radford2019language, yang2019xlnet, liu2019roberta}.
Fine-tuning using these architectures yields state-of-the-art results in the order of a few hours.
The caveat to these models is that the initial training can be on the scale of many days if not weeks, distributed across multiple GPUs \cite{strubell2019energy}, a costly endeavour.

\begin{figure}[!h]
    \centering
    \includegraphics[scale=0.0575]{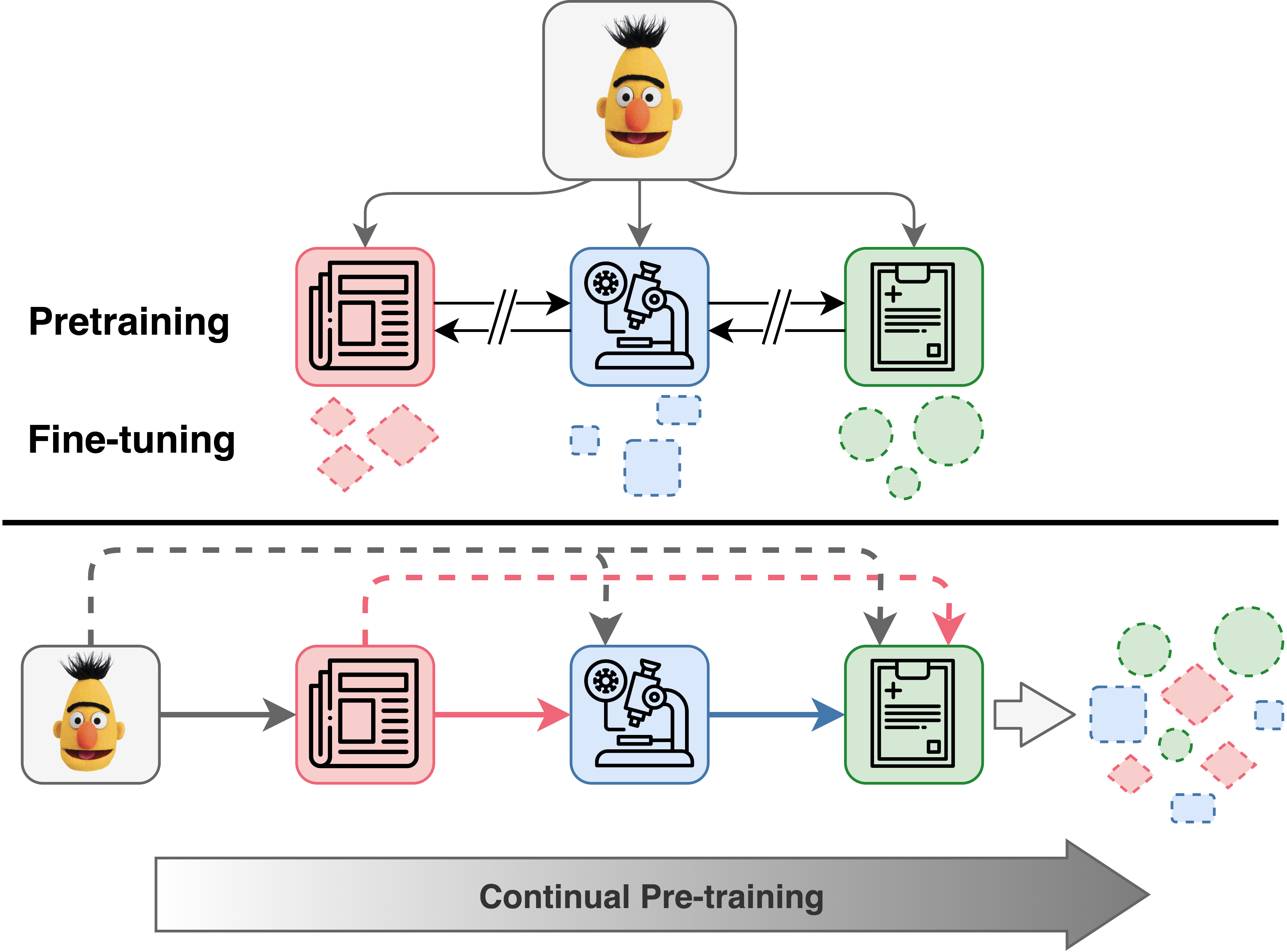}
    \centering
    \caption{Traditional approaches (top) train independent domain specific language models (newswire, bio-medical, and clinical) which share no cross domain knowledge.  They are further fine-tuned on their respective in-domain tasks. Our approach (bottom) shows how several domains are introduced in sequence, with knowledge retention using mitigation techniques across all domains.  Here the final model has the capability to properly fine-tune on any domain specific task.}
    \label{fig:model}
\end{figure}

Pre-trained language models are adapted to perform strongly in more specific domains as well.
For example, while the original BERT models \citep{devlin2018bert} were trained on English Wikipedia articles and BooksCorpus \citep{zhu2015aligning}, the same masked language modeling was continued on bio-medical data.
BioBERT \citep{lee2019biobert} was trained using Pubmed abstracts and full articles, meanwhile Clinical BERT \citep{alsentzer2019publicly} was further refined using MIMIC-III clinical notes \citep{johnson2016mimic}.
Evidence suggest that understanding the syntactic structure of scientific literature and clinical data from pre-training boosts performance in their respective downstream tasks \cite{peng2019transfer}.
Pre-training is performed with the expectation of building robust, high capacity generalized language models which continue to absorb new domain knowledge.

% Catastrophic forgetting  is the unfortunate side-effect of incorporating new domain data in a sequential manner. 
Unfortunately, continual learning \cite{Ring97child} suffers from catastrophic forgetting \citep{mccloskey1989catastrophic, ratcliff1990connectionist} when incorporating domain data in a sequential manner.
Parameters shift towards capturing the current task (or domain) and if previous data is no longer available the model will lose representation of it.
% In general perplexity increases for older domains, and models lose confidence in continual learning settings \cite{yogatama2019learning}.
For many tasks the straightforward solution is to combine datasets during training and approach this as a multi-task learning (MTL) \citep{ruder2017overview} problem.
Mixing data
has the desired effect of constraining parameters to find a space where both tasks reach close to optimal performance.

We argue that these expensive pre-trained models are an example where MTL is not feasible in practice for several reasons.
Time and hardware accessibility are the largest constraints for developing such systems.
Access to large scale training data is generally not possible \citep{radford2019language, devlin2018bert}, and exact training configurations are equally difficult to gather with results being arduous to reproduce.
Resource usage has recently been criticized from another perspective as well.
\citet{strubell2019energy} show that as deep neural architectures in the natural language community grow we increasingly trade results for carbon emissions. 

Our work conducts an empirical investigation into suitable methods for multi-domain pre-training in a continual learning setting.
We focus our efforts towards three methods: (i) elastic weight consolidation (EWC), (ii) learning rate control (LRC), and (iii) experience replay (ER).
EWC \cite{kirkpatrick2017overcoming} is a parameter constraining method, an upgrade to vanilla regularization (e.g. $L_2$).
LRC is borrowed from stage two of ULMFiT \cite{howard2018universal} pre-training as a data independent method.
Finally, as a scaled back version of MTL we investigate experience replay (ER), re-introducing data at a fixed scale from previous domains during pre-training.
Furthermore we explore data selection approaches to improve efficiency for both ER, and EWC.

Our goal is to understand the trade-offs across these models in terms of resources and setup.  
To this end we conduct experiments across multiple domain shifts while pre-training.
To evaluate the efficacy of the methods we use downstream fine-tuning tasks in the domains we study.
To better understand how knowledge across domains is transferred, we perform layer-wise analysis and observe that outer layer are the most transferable. 
% This indicates that we can train using these penultimate layers, which is efficient especially for fine-tuning large LMs.

%We find that we can efficiently fine-tune just the

% A solution, and a scaled back version of MTL is experience replay (ER).
% The goal with ER is to select replay buffer to re-introduce previous samples back to the model as it trains on a new task or in a new domain.
% Due to the scale of data for pre-training even a trivial percentage can be on the order of millions of training samples\footnote{E.g. 1\% of RoBERTa pre-training data corresponds to $>400,000$ samples.}.
% At such numbers there is likely a large redundancy in small replay splits, and discovering heuristics to better data selection becomes prized.
% Our goal becomes to further reduce the cost of pre-training by exploring several signals to extend replay buffer coverage while keeping the number of samples to a minimum.

% Constraint based methods are also popular to mitigate catastrophic learning, therefore we explore two such paradigms in the form of elastic weight consolidation (EWC) and learning rate control (LRC).
% Finally, it is not well understood how large transformer based models understand input at each layer, we seek to fill in gaps of model interpretability to aid us in discovering better paradigms for continual pre-training.

Our contributions are as follows \footnote{Our code is avaialble at \url{https://github.com/aws-health-ai/multi_domain_lm}}:
\begin{itemize}
    \item
    We provide empirical evidence of catastrophic forgetting mitigation with experience replay, learning rate control, and elastic weight consolidation, applied towards large scale language model pre-training.  To this we add multiple domain shifts into bio-medical, and clinical data.
    \item
    We explore various data selection approaches for both elastic weight consolidation and replay based models.
    \item
    We investigate layer-wise understanding for continual pre-training across several domains to understand how best to mitigate forgetting and transfer knowledge understanding.
\end{itemize}

\begin{table*}
    \centering
    \resizebox{\textwidth}{!}{
    \begin{tabular}{l|rrrrrrrrr}
    Model & CoLA & SST-2 & MRPC & STS-B & QQP & MNLI & QNLI & RTE & WNLI  \\
    \toprule
    BERT\textsubscript{BASE} & 57.82 & 92.09 & 86.74 & 88.13 & 87.49 & 84.01 & 90.79 & 64.98 & 53.52\\
    BioBERT & 37.78 & 89.68 & 88.44 & 87.40 & 86.96 & 83.19 & 89.79 & 60.29 & 28.17 \\
    \midrule
    Delta & 20.04 & 2.41 & -1.69 & 0.73 & 0.53 & 0.82 & 1.01 & 4.69 & 25.35  \\
    \bottomrule
    \end{tabular}}
    \caption{Performance drop of BioBERT after further pre-training on Pubmed articles.  The last row shows a positive value indicating the degree to which performance has dropped, and a negative value when it has increased.}
    \label{table:motivation}
\end{table*}

\section{Continual Learning}

We empirically study three forms of mitigation for catastrophic forgetting.
Constraint based training in the form of EWC and learning rate control, and experience replay.

\subsection{Elastic Weight Consolidation}

EWC makes use of a simple Bayesian factorization of model representation \cite{kirkpatrick2017overcoming}.
This isolates the posterior of a learned task (A) while maintaining the objective of a current task (B).
Due to the intractability of the true posterior, EWC makes use of a Fisher information \citep{frieden2004science} matrix diagonal to approximate the effect of Task A on the parameters of a model.
Intuitively speaking, if a parameter had a large effect on task A the Fisher value would be small yielding low variance to adapt to task B.
This holds true inversely for when the Fisher value is large.

In practice, we initialize the Fisher matrix using gradients calculated with data sampled from Task A, which has already converged \cite{spall2005monte}.
This is demonstrated in Eq. 1 where $i$ and $j$ index parameters and data samples respectively.
\begin{align}
    F_{i,i} &= \frac{1}{N} \sum_{j=1}^{N} \Big( \frac{\partial \mathcal{L}_A^{(j)}}{\partial \theta_i} \Big)^2 \\
    \mathcal{L}(\theta) &= \mathcal{L}_B(\theta) + \sum_i \lambda F_{i,i}(\theta_i - \theta_{A,i}^*)^2
\end{align}
The full objective for task B is given in Eq. 2 where $\mathcal{L}_B(\theta)$ is the loss function of Task B, and EWC is represented as the second term regularizing model parameters.
Specifically by weighting the shift of model parameters while training on Task B (here $\theta_i$ and $\theta_{A,i}^*$ being the currently updated and frozen Task A parameters at index $i$ respectively).
The EWC objective component is further adjusted by the hyperparameter $\lambda$.

\subsection{Learning rate control}
\label{section:LRC}

Our approach models the second stage of ULMFiT \cite{howard2018universal}, namely target task fine-tuning.
We begin with a layer wise modifications by applying a decaying learning rate as a function of layer depth moving from the last layer towards model input.
\begin{align}
\eta^{(l-1)} = \frac{\eta^{(l)}}{\rho}
\end{align}
Here $\eta$, $l$, and $\rho$ denote learning rate, layer index and decay rate respectively.
Depth plays a factor in our model since the network consists of 14 layers (i.e. 12 transformer layers, one layer for input, and one for the LM head).
% Additionally, we switch from the polynomial decay learning rate scheduler to slanted triangular learning rate (STLR).

\subsection{Experience Replay}
Given a replay buffer of a fixed, limited size we empirically investigate sample efficiency over a number of heuristic data selection methods.
We focus our attention on how best to select data for this buffer, hypothesizing that domain coverage will increase performance.
Recent work \cite{d2019episodic} has shown how this is crucial in strict lifelong learning when updating a fixed buffer size.
% We explore features generated from prior models to determine if heuristic data selection is useful for replay based models.

\section{Catastrophic Forgetting in Language Modeling}
\label{section:catastrophic_forgetting}
We motivate our own experiments by first exploring off-the-shelf models to get a sense of the problem.
To this end we fine tuned a BERT\textsubscript{BASE} architecture on all nine GLUE \cite{wang2018glue} tasks.
These were compared directly against BioBERT, which has been further trained on full Pubmed articles.
As reported in Table \ref{table:motivation} an overall trend of performance deterioration is apparent with a relative increased error of $7.64 \%$ in the bio-medical model.
% BioBERT performed negligibly better than original BERT on only a single task (MRPC).
Furthermore, we observed that on tasks which BERT struggles with, such as CoLA and WNLI, the performance decrease is amplified when switching pre-training domains.
% We are therefore  motivated not only to mitigate the negative trend in generic performance, but also that some form of backward transfer may increase the numbers in a previous domain.

\section{Experimental Details}
\label{section:details}
We first cover the data domains, fine-tuning tasks, and general modeling setup used in both our heuristic search as well as our main experiments in Section \ref{section:stage_1}.

\subsection{Pre-training Data}
We processed publicly available bio-medical and non-bio-medical corpora for pre-training our models.
For non-bio-medical data, we use BookCorpus and English Wikipedia data, CommonCrawl Stories \cite{trinh2018simple}, and OpenWebText \cite{Gokaslan2019OpenWeb}.
This combined corpus contains roughly 18B tokens.
For bio-medical data, we use full Pubmed\footnote{https://www.ncbi.nlm.nih.gov/pmc/} articles which we processed to remove all tables, references, equations, and figures.
This yields a dataset of over 4B tokens.
For all datasets we retain training, validation, and test splits sampled at the document level with a respective ratio of 8:1:1.

\subsection{Evaluation Data}
% To evaluate modeling we track the perplexity of held-out test data for both domains.
We report the average accuracy across GLUE \cite{wang2018glue} tasks to track the performance of the model on generic natural language understanding.
For measuring performance on GLUE, we further limit the selection of tasks to be the five most deteriorated (i.e. CoLA \cite{cola}, SST-2 \cite{sst}, MNLI \cite{mnli}, QNLI \cite{qnli} and RTE \cite{rte}).
Tasks such as QQP\footnote{https://www.quora.com/q/quoradata/First-Quora-Dataset-Release-Question-Pairs} and MRPC \cite{mrpc} are generally robust against domain change and perform well regardless of initialization.
These five tasks reflect our findings from Table \ref{table:motivation}.
Additionally we evaluate on CoNLL-03 \cite{sang2003introduction} named entity recognition (NER), and SQuAD 1.1 \cite{qnli} question answering (QA).
To demonstrate domain shift we evaluate using BC5CDR \cite{li2016biocreative}, Chemprot \cite{krallinger2017overview} and BioASQ \cite{nentidis2019results} which are bio-medical NER, relation extraction (RE), and QA tasks respectively.
The first dataset is from the 2015 CDR challenge for identifying chemicals and diseases expertly annotated from Pubmed abstracts \footnote{We used a combined dataset: \url{https://github.com/cambridgeltl/bmip-2018}.}.
Chemprot contains annotations of chemical-protein reactions, also taken from Pubmed articles.
Finally BioASQ appears in our paper using the same format and splits as described by \citet{blurb}.
Namely QA is treated as a binary classification of whether the answer to the query exists in the provided context.

\subsection{Modeling}
For modeling we use the RoBERTa architecture \cite{liu2019roberta}, and implement EWC, learning rate control, and experience replay changes directly into the model\footnote{\url{https://github.com/pytorch/fairseq/tree/master/examples/roberta}}.
This extension of the original BERT removed next sentence prediction and is trained using only masked language modeling using very large batch sizes.
We utilize all training hyperparameters as provided by \citet{liu2019roberta}
unless otherwise noted, and use RoBERTa \textsubscript{BASE} as parameter initialization for all experiments.
As a form of deterioration understanding, we continue to train a model using Pubmed articles (denoted as PMC) with no mitigation techniques.

% \begin{figure}[t]
%     \centering
%     \includegraphics[scale=0.5]{gradient_distribution.png}
%     \centering
%     \caption{Distribution of gradients using 400,000 samples from the RoBERTa pre-training corpus.  For ease of viewing we limit the left edge to $1.0e^{7}$.}
%     \label{fig:grad_dist}
% \end{figure}

% Since we discuss data and modeling in this section, it may be best moved until after data + modeling is explained
\section{Data Selection Methods}
\label{section:heuristics}
Data selection is an important component of both supervised, and unsupervised training.
In our case, there is an abundance of data to build both the Fisher matrix, as well as the replay buffer.
To do this efficiently for EWC and ER we need to severely restrict the number of datapoints we utilize.
For example a mere $1.0\%$ of generic pre-training data makes up over 400k segments.
We require this subset to be comprehensively representative of the domain.
Therefore, rather than randomly sampling data, we can use model generated features to induce better coverage of previous domains.

\begin{table}[t]
    \centering
    \resizebox{\columnwidth}{!}{
    \begin{tabular}{lrrr}
    \toprule
    Sampling Type & GLUE & SQuAD & Avg.\\
    \midrule
    RoBERTa \textsubscript{BASE} & 87.56 & 90.20 &  88.00\\
    RoBERTa \textsubscript{PMC} & 83.00 & 88.73 & 83.95 \\
    \midrule
    \multirow{4}{*}{\rotatebox{90}{\textsc{ER}}} \quad Random & 84.23 & 89.43 & 85.10 \\
    \quad\quad High & 84.59 & 87.99 & 85.15 \\
    \quad\quad Low & 83.99 & 88.97 & 84.82 \\
    \quad\quad Uniform & 84.69 & 89.70 & \textbf{85.53} \\
    \midrule
    \multirow{4}{*}{\rotatebox{90}{\textsc{EWC}}} \quad Random & 86.93 & 90.32 & 87.50 \\
    \quad\quad High & 87.08 & 90.27 & \textbf{87.61} \\
    \quad\quad Low & 86.64 & 90.49 & 87.28 \\
    \quad\quad Uniform & 87.03 & 90.43 & 87.60 \\ 
    \bottomrule
    \end{tabular}}
    \caption{Four sampling techniques used for pre-training and evaluated on GLUE and SQuAD 1.1. The results are compared against RoBERTa \textsubscript{BASE} and an unmitigated model trained on Pubmed articles (denoted using \textsubscript{PMC}).   The average column takes into account each of the individual GLUE tasks.}
    \label{table:sampling_grad}
\end{table}

\subsection{Gradient Analysis}
\label{section:grad_analysis}
We begin by treating the sum of squared gradients as a one-dimensional feature for data selection.
The generic data is a skewed distribution with a mean at $1.04e^{7}$ and a standard deviation and max values of $4.89e^{8}$, and $1.82e^{11}$ respectively.
The lower bound is, of course, $0$ and arguably the samples closer towards that bound are more representative of the model in its generic state given this long tail.

To be thorough we sampled data from this domain in four different ways: (i) randomly, (ii) low, (iii) high, and (iv) uniformly.
For low and high sampling we order the samples according to this feature value and slice the list from the front or back.
For uniform sampling we bin the data according to the gradient value, and sample from the bins uniformly, whereas random sampling is performed by treating all samples equally.
For each of these experiments we sample $0.1\%$ of the total corpus (roughly $42$k segments).
Furthermore in the same way that ER uses data to construct the replay buffer, EWC uses the samples to build the Fisher diagonal.
We therefore test each sampling method across both mitigation techniques.

% \begin{table*}[ht]
%     \centering
%     \begin{tabular}{rrrrr|rrr}
%     \toprule
%      & & \multicolumn{3}{c}{ER} & \multicolumn{3}{c}{EWC} \\
%      \cmidrule(lr){3-5}\cmidrule(lr){6-8}
%     \quad PCA & GMM & GLUE & SQuAD & Avg. & GLUE & SQuAD & Avg.\\
%     \midrule
%     \multirow{4}{*}{\rotatebox{90}{{\fontfamily{qcr}\selectfont <s>}}} \quad 50 & 5 & 84.18 & 89.33 & 85.04 & 86.86 & 90.46 & 87.46\\
%     \quad\quad 50 & 10 & 84.99 & 89.10 & 85.67 & 86.68 & 90.07 & 87.25 \\
%     \quad\quad 100 & 5 & 84.72 & 89.19 & 85.46 & 87.08 & 90.22 & \textbf{87.61}\\
%     \quad\quad 100 & 10 & 85.03 & 89.31 & \textbf{85.74} & 86.65 & 90.43 & 87.28\\
%     \midrule
%     \multirow{4}{*}{\rotatebox{90}{\textsc{Avg. Pool}}} \quad 50 & 5 & 84.22 & 89.31 & 85.06 & 86.67 & 90.10 & 87.24 \\
%     \quad\quad 50 & 10 & 84.17 & 89.38 & 85.04 & 86.55 & 90.43 & 87.20\\
%     \quad\quad 100 & 5 & 84.26 & 88.50 & 84.96 & 87.32 & 90.33 & \textbf{87.83}\\
%     \quad\quad 100 & 10 & 84.64 & 89.17 & \textbf{85.39} & 86.62 & 90.34 & 87.24\\
%     \bottomrule
%     \end{tabular}
%     \caption{}
%     \label{table:sampling_latent}
% \end{table*}

To test the effectiveness of our methods we pre-train RoBERTa \textsubscript{BASE} on one epoch of Pubmed data (with and without mitigation) and test retention performance by fine-tuning our models across GLUE and SQuAD 1.1.
Looking at Table \ref{table:sampling_grad} we see that above all, using low gradients is the least useful signal.
For ER, using uniform rather than low value selection has an average performance increase of $0.71$ points.
The other methods fall in line with uniform sampling indicating that including samples with larger gradients is helpful in representing of the source domain.
EWC appears to be more robust to data sampling with lower variance ($1.8\mathrm{e}{-2}$ vs. $6.4\mathrm{e}{-2}$ for ER) across all models, with high and uniform selection improving most.
% Incidentally samples with small or large gradient values were inconsistent leading us to believe that uniform sampling provides the best coverage when pre-training using such a skewed distribution.

\subsection{Sampling Latent Clusters}
We further investigate more feature-rich representations in the form of sentence embeddings.
\citet{aharoni2020unsupervised} have demonstrated that transformer based LMs exhibit a keen ability to distinguish domains via clustering.
The pre-training data for RoBERTa also comes from a variety of sources, with variation in prose, diction, and formality.
We therefore cluster this data to see both how it is distributed and if uniformly sampling from these groups yields good performance for both EWC and ER.

\citet{aharoni2020unsupervised} used average pooling across the last encoder layer to represent each segment, we test this method against using the vector representation of {\fontfamily{qcr}\selectfont <s>} ({\fontfamily{qcr}\selectfont [CLS]} in BERT) since it is frequently used in practice for sentence labeling.
We then use PCA \cite{WOLD198737} to reduce the dimensionality to $d \in \{50, 100\}$ and apply a Gaussian Mixture Model \cite{Reynolds2009} using $k \in\{5, 10\}$ as the number of clusters.

\begin{table}[t]
    \centering
    \resizebox{\columnwidth}{!}{
    \begin{tabular}{rrr|r}
    \toprule
    \quad PCA & GMM  &  ER Avg. & EWC  Avg.\\
    \midrule
    \multirow{4}{*}{\rotatebox{90}{{\fontfamily{qcr}\selectfont <s>}}} \quad 50 & 5 &  85.04 & 87.46\\
    \quad\quad 50 & 10 & 85.67 & 87.25 \\
    \quad\quad 100 & 5 &  85.46  & \textbf{87.61}\\
    \quad\quad 100 & 10  & \textbf{85.74}  & 87.28\\
    \midrule
    \multirow{4}{*}{\rotatebox{90}{\textsc{Avg. Pool}}} \quad 50 & 5 & 85.06  & 87.24 \\
    \quad\quad 50 & 10  & 85.04 & 87.20\\
    \quad\quad 100 & 5  & 84.96 & \textbf{87.83}\\
    \quad\quad 100 & 10  & \textbf{85.39} & 87.24\\
    \bottomrule
    \end{tabular}}
    \caption{GLUE and SQuAD average performance for both ER and EWC when using two pooling techniques.}
    \label{table:sampling_latent}
\end{table}

The resulting experiments for both ER and EWC can be seen in Table \ref{table:sampling_latent}.
Using PCA at $100$ provides higher metrics for both ER and EWC, while the number of clusters for GMM does not give an interpretable signal across the experiments.

We note that from a practical perspective it is much faster to process data using clustering than gradients, largely due to the ability to batch data for clustering.
% On the other hand once gradient values are calculated and stored sampling from them takes a trivial amount of time.
Accumulating gradients for 1MM samples takes roughly five days using an NVIDIA V100, whereas acquiring latent representations from the same amount of data finishes in less than four hours (this does not account for PCA and clustering which takes an additional four to five hours).

\section{Mitigation of Catastrophic Forgetting}
We provide results for one and two stage domain shifts as given by fine-tuning tasks.
Again, we apply mitigation only to pre-training and express our model performance by using them to fine-tune downstream tasks.

\subsection{Setup}
For a baseline and potential upper bound of performance we train a multi-domain learning (denoted as MDL) model which utilizes the full combined generic and bio-medical training sets as input data.
For EWC (+EWC) we tune both $\lambda$ [$0.5$, \underline{$1.0$}, $5.0$, $10.0$], and the size of the data used for fisher initialization [\underline{$0.1\%$}, $1.0\%$, $10.0\%$]; best values are underlined.
For experience replay (+ER) we experiment with mixing non-bio-medical data (the same  subset used for EWC init.) in each batch with a ratio proportional to their sizes.
Additionally we showcase both a gradient based sampling (denoted with a subscript unif), and the GMM-PCA (subscript GMM)  ($k=5$, $d=100$) for both ER and EWC.
We tuned the decay rate, $\rho$ in Eq. 3 [\underline{$1.3$}, $1.7$, $2.6$] for LRC.

\begin{table*}[ht]
    \centering
    \resizebox{\textwidth}{!}{
    \begin{tabular}{lrrrr|rrrr|r}
    \toprule
    &  \multicolumn{4}{c|}{generic}   &  \multicolumn{4}{c|}{bio-medical}  & \\
    Model & GLUE & SQuAD & CoNLL & Avg. & BC5CDR & Chemprot & BioASQ & Avg. & Overall  \\
    \midrule
    RoBERTa \textsubscript{BASE} & 87.56 & 90.20 & 90.11 & 88.30 & 84.94 & 63.27 & 75.41 & 74.69 & 81.49 \\
    PMC & 83.00 & 88.73 & 87.35 & 84.44 & 86.68 & 65.13 & 75.41 & 75.74 & 80.09 \\
    \midrule
    MDL & 84.89 & 88.92 & 89.72 & 86.15 & 85.76 & 65.16 & 75.41 & 75.44 & 80.79\\
    PMC +LRC & 86.78 & 90.35 & 89.76  & 87.72 & 85.47 & 62.30 & 75.41 & 74.39 & 81.05\\
    PMC +ER\textsubscript{unif} & 84.69 & 89.70 & 89.10 & 86.04 & \textbf{87.20} & \textbf{67.40} & 77.13 & 77.24 & 81.64 \\
    PMC +ER\textsubscript{GMM} & 84.25 & 88.50 & 89.78 & 85.65 & 86.83 & 63.70 & \textbf{82.42} & \textbf{77.65} & 81.65 \\
    PMC +EWC\textsubscript{unif} & 87.03 & \textbf{90.43} & 89.77 & 87.90 & 86.23 & 65.90 & 79.73 & 77.28 & \textbf{82.59} \\
    PMC +EWC\textsubscript{GMM} & \textbf{87.08} & 90.22 & \textbf{90.46} & \textbf{88.01} & 86.05 & 65.50 & 76.18 & 75.90 &  81.96\\
    \bottomrule
    \end{tabular}}
    \caption{Single stage domain adaptation.  Other than RoBERTa \textsubscript{BASE}, each model is pre-trained further on one epoch of bio-medical data.  We average generic performance across five GLUE tasks, as well as QA (from SQuAD), and NER (CoNLL). The average across generic tasks considers all nine tasks equally. bio-medical performance is for BC5CDR (NER), Chemprot (RE), and BioASQ (QA) with the overall performance being the mean for bio-medical and generic averages.}
    \label{table:results_main}
\end{table*}

\subsection{Results}
Our experimental results are reported in Table \ref{table:results_main}.
The first two rows contain the off-the-shelf RoBERTa as well as the  PMC setting which received no catastrophic forgetting mitigation when further trained on bio-medical data.
The lower section lists all mitigation based experimental settings as described above.
For all models pre-trained using Pubmed data we fine-tune on tasks after a single epoch of pre-training.

We divide columns by task domain.
The first three tasks (i.e. GLUE, SQuAD, and CoNLL) cover generic domain understanding.
Just as in Section \ref{section:grad_analysis} we use the five worst GLUE tasks.
For an overall understanding of forgetting we provide the average across all generic tasks.
bio-medical tasks are displayed next followed by overall performance weighing the bio-medical and generic tasks equally \footnote{We take the mean of the generic and bio-medical average rather than treating each task equally since there are significantly more generic tasks.}.
NER and RE scores are reported using micro-$F_1$; all GLUE tasks we report accuracy on the development set; SQuAD is evaluated using $F_1$; BioASQ uses accuracy.

\subsubsection{Catastrophic Forgetting}
Unsurprisingly among the first two rows RoBERTa \textsubscript{BASE} performs best overall on generic tasks with an average performance increase of $4.47\%$ over the unmitigated (PMC) model.
Conversely it under-performs on the bio-medical tasks, validating the need to further pre-train on domain specific data.
When averaging across the three bio-medical tasks the PMC model has a $1.05$ point $F_1$ edge.
It should be noted here that four of the models achieved the same BioASQ $F_1$ score, this was not reported in error.

\subsubsection{Mitigation Based Models}
\label{section:stage_1}
EWC and LRC both respond well during domain shifts, are our best candidates for combating catastrophic forgetting, and average only half a point in deterioration amongst the three of them when compared against RoBERTa \textsubscript{BASE}.
% Furthermore this model has the highest combined confidence when observing perplexity across domains.
LRC has the benefit of tuning a single hyperparameter, the decay rate ($\rho$).
Due to the depth of the models we found that a high value ($\rho=2.6$) yields a model which has a negligible drop in performance for generic tasks (with an average of $88.28$) but had a more difficult time with later domains.

We observed during hyper-parameter optimization that EWC was quite sensitive to $\lambda$ values.
With higher coefficients ($\lambda > 1.0$) EWC was able to halt deterioration nearly completely but performed quite poorly on bio-medical tasks.
To better understand the importance of the Fisher values, we trained EWC with no Fisher (i.e removing $F_{i,i}$ from Eq. 2). 
We found that this resulted in less competitive bio-medical results (averaging $3.68\%$ worse than the listed bio-medical EWC scores, and having overall the worst scores for both bio-medical tasks across all models), illustrating that giving equal weight to all the parameters results in poor generalization across source and target domains.
MDL performed surprisingly average compared to the resource trade-off of the model.
While it does produce better results than RoBERTa \textsubscript{BASE} in the bio-medical domain, the model struggles to retain generic knowledge.
Experience replay grapples most with domain retention and produced the highest mitigated BC5CDR, Chemprot, and BioASQ results coupled with the lowest generic results.

When comparing sampling techniques across a larger number of fine-tuning experiments we echo results from Section \ref{section:heuristics}.
Experience replay is stronger when using gradient based sampling, while EWC functions better using clustered latent representations.
Therefore, in practice, we would suggest latent representations for better efficiency.

\begin{table*}[ht]
    \centering
    \resizebox{\textwidth}{!}{
    \begin{tabular}{lrr|rrrr|r}
    \toprule
    Model & Generic & bio-medical & i2b2 NER & i2b2 RE & ADE RE & Clin. Avg. & Overall\\
    \midrule
        RoBERTa \textsubscript{BASE} & 88.30 & 74.69 & 81.12 & 77.16 & 87.82 & 82.03 & 81.67\\
        PMC, clin. & 82.98 & 76.53 & 85.96 & 79.44 & 88.96 & 84.79 & 81.43\\
        \midrule
        LRC\textsubscript{2}  & \textbf{87.47} & 74.33 & 85.03 & 77.93 & 86.84 & 83.26 & 81.69\\
        ER\textsubscript{2}  & 84.51 & \textbf{75.85} & 85.16 & 79.20 & \textbf{88.23} & \textbf{84.20} & 81.52\\
        EWC\textsubscript{2}  & 86.99 & 75.04 & \textbf{85.43} & \textbf{79.59} & 86.07 & 83.47 & \textbf{81.91}\\
    \bottomrule
    \end{tabular}}
    \caption{Averaged performance for all generic, and bio-medical tasks (i.e. as seen in Table \ref{table:results_main}).  Clinical average is across i2b2 NER and RE as well as n2c2 ADE RE are given as Micro-$F_1$}
    \label{table:two_stage}
\end{table*}

\subsubsection{Two Stage Domain Adaptation}
To further evaluate mitigation methods we continue pre-training models using clinical data.
We chose the clinical domain since although it may appear close to bio-medical text, health records have been shown to differ drastically in prose and diction even when the underlying information may be similar \cite{blurb}.
We processed 659M tokens of de-identified clinical notes and continued training using the PMC +LRC, PMC +ER \textsubscript{unif}, and PMC +EWC \textsubscript{GMM} from Table \ref{table:results_main} (with this stage of model denoted with a subscript 2).
RoBERTa \textsubscript{BASE} is the untouched model as presented in Table \ref{table:results_main}, and we continue to train (unmitigated) the PMC model from the same table (now denoted as PMC, clin.).
We evaluate models on RE and NER from the i2b2 challenge after 5 epochs \footnote{To determine an appropriate stopping point we evaluated each epoch using the the clinical NER task until the Micro-F$_1$ plateaued.}.
Additionally we use the n2c2 adverse drug reaction (ADE) \cite{n2c2} RE task.

Stage two results are reported in Table \ref{table:two_stage}.
The last column in this table indicates that average overall performance is about the same across models, however, when we take a closer look at the domain breakdown we see this is not the case.
As expected the unmitigated model (PMC, clin.) suffers from performance deterioration in generic tasks, with GLUE dropping drastically (an error increase to $6.21\%$ compared to RoBERTa \textsubscript{BASE}).
We find that LRC is still firmly holding onto generic representation, with the smallest drop in average generic performance of $0.83$ points, when compared to stage one.
Here we found that tuning $\rho$ became more prevalent, with the range of average clinical scores for LRC being $1.49$ points.
ER, and EWC are the only mitigated models which achieve competitive numbers for clinical tasks, although they both show a drop in generic, and bio-medical results.
Both of the latter models outperform the base model in average bio-medical and clinical metrics.

\section{Analysis}
To further understand learning and forgetting across different mitigation strategies, we conduct analyses to investigate how different layers of the model adapt to in-domain pre-training, whether the adaptation helps in transferring knowledge to downstream tasks, and how knowledge learned from in \& out of domain data cooperates together. 

\subsection{Layer-wise analyses}
\subsubsection{Weight Similarity} 
\label{sec:weightsimilarty}

\begin{figure*}[ht]
\centering
\subfloat[Generic vs. bio-medical]{\includegraphics[trim=0pt 0pt  0pt 0pt, clip=true, width=0.33\textwidth]{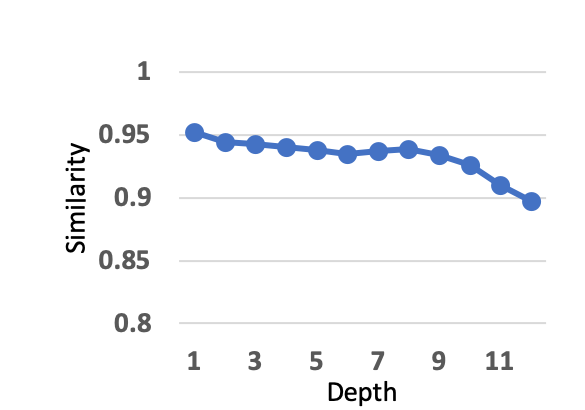}\label{fig:d0}}
\subfloat[Mitigated Models vs. Generic]{\includegraphics[trim=0pt 0pt  0pt 0pt, clip=true, width=0.33\textwidth]{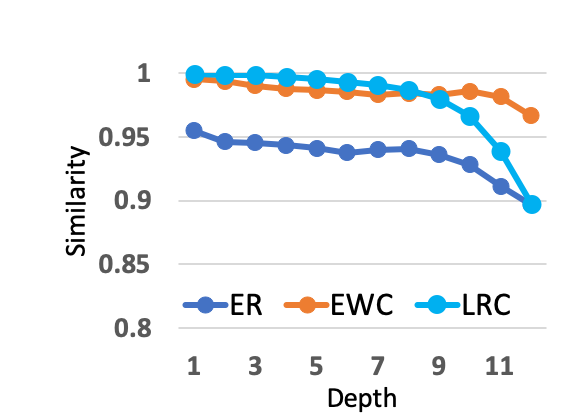}\label{fig:d1}}
\subfloat[Mitigated Models vs. bio-medical]{\includegraphics[trim=0pt 0pt  0pt 0pt, clip=true, width=0.33\textwidth]{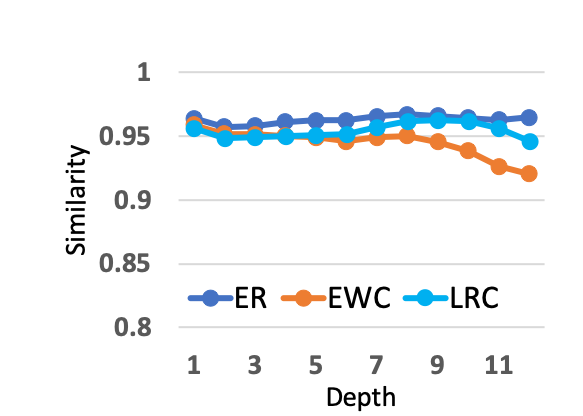}\label{fig:d2}}
\caption{Weight distance vs. Depth across two domains.  We compare RoBERTa \textsubscript{BASE} (trained on \textit{generic} data) against PMC (denoted as bio-medical) and two mitigated models. Distance is given using cosine similarity.}
\label{fig:distance}

\end{figure*}

Figure \ref{fig:distance} displays layer-wise weight (cosine-) similarity between models before and after pre-training on bio-medical data. 
We compare RoBERTa \textsubscript{BASE} (denoted as Generic) against the PMC model (row 2 in Table \ref{table:results_main} and denoted as bio-medical in the Figure).
In Figure \ref{fig:d0} we discern similarity in layers closer towards the input.
%In Figure \ref{fig:d0} we discern similarity in layers closer towards the input, indicating that grammatical understanding occurs in earlier layers, whereas segment level domain specific perception (i.e. semantics) appears in later layers, which is consistent with works from \citet{belinkov2018evaluating,jawahar2019does}.
By comparing Figures \ref{fig:d1} and \ref{fig:d2} which illustrate how mitigated models behave compared to one another, we find that ER allows the model parameters to shift much closer towards the bio-medical data while EWC finds a shared space for parameters in both models.
This is consistent with what we have observed in Section \ref{section:stage_1} where we find EWC is better at mitigating catastrophic forgetting compared to ER.
It was important to see how LRC weights behave as well.
Intuitively since the learning rate is close to $0$ near the model input, these layers will change very little.  
This is indeed the case with only the last layer showing significant shift.
% After fine-tuning on target domain (biomed), models pre-trained via ER are more closer to target domain (biomed) while models pre-trained via EWC are more closer to source domain (generic), as shown in Fig.\ref{fig:d1}, \ref{fig:d2}. 

We investigate if constraining the weights to a shared space is enough to produce a good overall model.
We observed that without the Fisher matrix, weight similarity between EWC and RoBERTa \textsubscript{BASE} is lower than $0.2$, which is confirmed by the low $F_1$ scores noted in Section \ref{section:stage_1}.
This indicates that the Fisher diagonal plays an important role in fluctuating variance.

\subsubsection{Transferability via Probing Tasks} 
\begin{figure}
\centering
\subfloat{\includegraphics[trim=20pt 0pt  0pt 30pt, clip=true, width=0.45\textwidth]{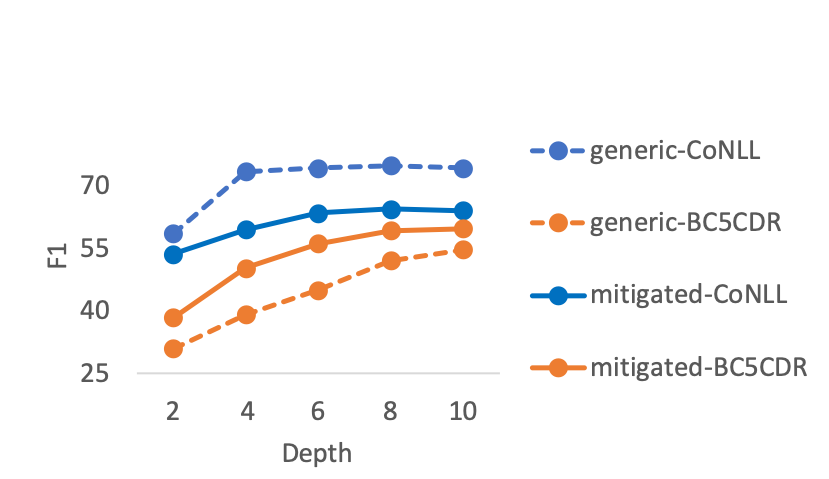}} 
\caption{Transferability vs. Depth. Dashed curves denote generic models and solid curves denote mitigated models. After fine-tuning on bio-medical data, the performance of CoNLL drops while the performance of BC5CDR is boosted.}
\label{fig:transfer}
\end{figure}

To evaluate layer-wise transferability of pre-trained LMs, we use NER as a probing task and limit the capacity of task-specific layers to focus on what information has been learned by the model. 
We evaluate each layer of pre-trained LMs by extracting the model output as features and only fine-tuning task-specific layers.
We observe in Figure \ref{fig:transfer} that (1) outer layers are most transferable to downstream tasks except for the last layer and (2) the performance of domain specific NER increases much faster than generic NER across layers, which indicates that grammatical understanding occurs in earlier layers, whereas segment level domain specific perception (i.e. semantics) appears in later layers. 
Both (1) and (2) are consistent with Figure \ref{fig:d0} where weights change more in outer layers. 
This trend was also observed in previous works \citet{belinkov2018evaluating,jawahar2019does}.
% This trend was also observed in the visio-linguistic domain by \cite{singh2020pre-training} where layer 11 was most transferable.

Base on layer-wise analyses in this section, we empirically find that the adaptation in outer layers plays a key role in mitigation, which suggests that a decaying learning rate as a function of layer depth is worth being incorporated into different mitigation strategies.

\subsection{Qualitative Examples}
We observe that CF mitigation techniques are able to assist in generalization on rare words by composing knowledge from both generic and bio-medical domains. In Figure \ref{figure:annot} (i) we observe that ``Norilsk'' occurs quite rarely in Newswire data, which is used for pre-training generic domain, however, it is frequent in Pubmed but size of pre-training data is small. 
Combining the two datasets in the form ER and EWC helps generalise the model understanding.
We provide additional examples of this phenomenon in Figure \ref{figure:annot} (ii) \& (iii).

\begin{figure*}[t]
\setlength{\tabcolsep}{8pt}
\renewcommand{\arraystretch}{1}
\centering
\begin{fontppl}
\resizebox{\textwidth}{!}{
\begin{tabular}{ll|llr}
\toprule
  & Text & Model & Label & conf. \\
\midrule
\textbf{(i)}:&Entire social infrastructures in the icy Far North where Norilsk & Ground Truth & {\fontfamily{qcr}\selectfont S-ORG} & --\\
            &is based depend on the company, and government has said that & RoBERTa \textsubscript{BASE} & {\fontfamily{qcr}\selectfont S-MISC} & 0.609\\ 
            &expenditure could far outstrip Norilsk 's debts. \textcolor{red}{\textbf{[Norilsk]}} & PMC & {\fontfamily{qcr}\selectfont S-MISC} & 0.983\\ 
            &officials declined to comment. & PMC+ER & {\fontfamily{qcr}\selectfont S-ORG} & 1.000\\ \\

\textbf{(ii)}:&President Arafat's position is clear that such a meeting should & Ground Truth & {\fontfamily{qcr}\selectfont S-LOC} & --\\ 
              &come after successful negotiations so that the meeting would have & RoBERTa \textsubscript{BASE} & {\fontfamily{qcr}\selectfont S-PER} & 0.998\\ 
              &positive results. Especially since the  \textcolor{red}{\textbf{[Hebron]}} issue has not been & PMC & {\fontfamily{qcr}\selectfont O} & 1.000\\ 
              &agreed yet and the crucial disputed issues have not been resolved.  & PMC+ER & {\fontfamily{qcr}\selectfont S-LOC} & 0.994\\ \\

\textbf{(iii)}:&The committee said the Italian club had violated regulations by & Ground Truth & {\fontfamily{qcr}\selectfont S-ORG} & --\\ 
                &failing to inform Feyenoord, with whom the player was under & RoBERTa \textsubscript{BASE} & {\fontfamily{qcr}\selectfont S-LOC} & 0.815\\ 
                &contract. Blinker was fined 75,000 Swiss francs (\$57,600) for & PMC & {\fontfamily{qcr}\selectfont S-LOC} & 1.000\\ 
                &failing to inform the English club of his previous commitment & PMC+ER & {\fontfamily{qcr}\selectfont S-ORG} & 1.000\\ 
                &to  \textcolor{red}{\textbf{[Udinese]}}.\\
\bottomrule
\end{tabular}}
\end{fontppl}
\caption{Multi-task effect: generalization of a model on rare words using shared knowledge of pre-training on Newswire and Pubmed data. Example spans (taken from the CoNLL test split) are passed through an NER system initialized with various pre-trained encoders.  We provide the labels and confidences for each.}
\label{figure:annot}
\end{figure*}

\section{Related Work}
Current work in catastrophic forgetting mitigation in NLP has been limited.
\citet{howard2018universal} introduced a multi stage training scheme for fine tuning LSTM based universal language models (ULMFiT).
The authors proposed that current methods, rather than data, are ineffective and focused on learning rate control across layers, as well as modifying learning rate scheduling.
A larger category of work deals with constraining model parameters to a latent space where they continue to capture previous tasks.
Initial work focused on model regularization and varying activations \cite{goodfellow2013empirical}. \citet{kirkpatrick2017overcoming} provided a more sophisticated solution constraining weights individually termed elastic weight consolidation (EWC).
We make use of both EWC and ULMFiT and provide further technical detail in this paper.
The final approach is focused on experience replay.
Using small samples of data from previous tasks coupled with local adaptation \citet{d2019episodic} demonstrate improvement in a lifelong learning training scheme.
\citet{chaudhry2019continual} also explore lifelong learning by experimenting with updating the memory bank for experience replay.
Our work focuses on both of these techniques with the major difference being problem scale.
Many existing works apply these solutions on small networks whereas we experiment on architectures having several orders of magnitude more parameters.

There has been a recent focus on more effective pre-training which focuses on narrowing the pre-training domain as we move closer towards fine-tuning.
STILTs \cite{stilts} and TandA \cite{garg2019tanda} use intermediate tasks (in a data rich domain) training to lower variance during target task fine-tuning.
This intuition was also covered in the visio-linguistic domain by \citet{singh2020pretraining}.
Finally \citet{gururangan2020dont} work on MLM pre-training and provide conclusive evidence at scale of the works listed above.
This last body of work, although dealing with pre-training is different from our work in that we study mitigation of domain forgetting, rather than reducing variance by adding intermediate domains or tasks to pre-training.
\vspace{-0.05in}
\section{Conclusion}
\vspace{-0.05in}
In this work, we empirically investigated the existence of catastrophic forgetting in large language model pre-training.
We further explored constraint and replay based mitigation techniques to close the performance gap between general and domain specific natural language tasks.
We find that training a single model across multiple domains is possible.
Due to practical considerations, we would suggest using latent representation for data selection when working with a data dependent model such as ER or EWC.
When no previous data is available LRC provides a simple yet powerful solution for retaining prior domain knowledge.
In the future work wish to explore more data independent methods such as LRC, for both speed and lack of data dependency, as well as manipulation of the decay w.r.t. what we have discovered from our layer-wise analysis.
\vspace{-0.05in}
\section{Acknowledgement}
\vspace{-0.05in}
We would like to thank Byron Wallace, Kevin Small, Ramesh Nallapati, and members of Amazon Comprehend Medical for their help in shaping this work over the last year, as well as the conference reviewers for providing thoughtful feedback.

\bibliography{main}
\bibliographystyle{acl_natbib}

\end{document}